# Summarization from medical documents: a survey

Stergos Afantenos♣, Vangelis Karkaletsis♣, Panagiotis Stamatopoulos♦

♣Software & Knowledge Engineering Laboratory
Institute of Informatics & Telecommunications
National Centre for Scientific Research (NCSR) "Demokritos"
15310 Aghia Paraskevi Attikis, Athens, Greece
Tel. +30-210-6503149, e-mail: {stergos, vangelis}@iit.demokritos.gr

♦Department of Informatics, University of Athens
TYPA Buildings, Panepistimiopolis
GR-157 71 Athens, Greece
Tel. +30-210-7752222, e-mail: T.Stamatopoulos@di.uoa.gr

## Abstract

**Objective:** The aim of this paper is to survey the recent work in medical documents summarization.
**Background:** During the last decade, documents summarization got increasing attention by the AI research community. More recently it also attracted the interest of the medical research community as well, due to the enormous growth of information that is available to the physicians and researchers in medicine, through the large and growing number of published journals, conference proceedings, medical sites and portals on the World Wide Web, electronic medical records, etc.
**Methodology:** This survey gives first a general background on documents summarization, presenting the factors that summarization depends upon, discussing evaluation issues and describing briefly the various types of summarization techniques. It then examines the characteristics of the medical domain through the different types of medical documents. Finally, it presents and discusses the summarization techniques used so far in the medical domain, referring to the corresponding systems and their characteristics.
**Discussion and Conclusions:** The paper discusses thoroughly the promising paths for future research in medical documents summarization. It mainly focuses on the issue of scaling to large collections of documents in various languages and from different media, on personalization issues, on portability to new sub-domains, and on the integration of summarization technology in practical applications.

## 1. Introduction

New technologies, such as high-speed networks and inexpensive massive storage, along with the remarkable growth of the Internet, have led to an enormous increase in the amount and availability of on-line documents. This is also the case for medical information, which is now available from a variety of sources. However, information is only valuable to the extent that it is accessible, easily retrieved and concerns the personal interests of the user. The growing volume of data, the lack of structured information, and the information diversity have made information and knowledge management a real challenge towards the effort to support the medical society. It has been realized that added value is not gained merely through larger quantities of data, but through easier access to the required information at the right time and in the most suitable form. Thus, there is a strong need for improved means of facilitating information access.

The medical domain suffers particularly from the problem of information overload since it is crucial for physicians and researchers in medicine and biology to have quick and efficient access to up-to-date information according to their interests and needs. Considering, for instance, the scientific medical articles, [1, p 38] state that: *"...there are five journals which publish papers in the narrow specialty for cardiac anesthesiology, but 35 different anesthesia journals in general; approximately 100 journals in the closely related fields of cardiology (60) and cardiothoracic surgery (40); and over 1000 journals in the more gen-*

*eral field of internal medicine."* The situation becomes much worse if one considers relevant journals or newsletters in other languages, Web sites with relevant information, medical reports, etc.

Given the number and diversity of medical information sources, methods must be found that will enable users to quickly assimilate and determine the content of a document. Summarization is one such approach that can help users to quickly determine the main points of a document. [2] provide the following definition for a summary: "*A summary can be loosely defined as a text that is produced from one or more texts, that conveys important information in the original text(s), and that is no longer than half of the original text(s) and usually significantly less than that. Text here is used rather loosely and can refer to speech, multimedia documents, hypertext, etc. The main goal of a summary is to present the main ideas in a document in less space. If all sentences in a text document were of equal importance, producing a summary would not be very effective, as any reduction in the size of a document would carry a proportional decrease in its informativeness. Luckily, information content in a document appears in bursts, and one can therefore distinguish between more and less informative segments. Identifying the informative segments at the expense of the rest is the main challenge in summarization.*"

Although initial work on Summarization dates back to the late 50's and 60's (*e.g.* [3,4]), followed by some sparse publications (*e.g.* [5,6]), most research in the field has been carried out during the last decade. During these last few years, researchers examined a great variety of techniques and applied them in different domains and genres of documents, in order to see which are the ones that yield the most practical results for each domain and genre.

This survey presents the potential of summarization technology in the medical domain, based on the examination of the state of the art, as well as of existing medical document types and summarization applications. An important aspect of this survey is that it is not restricted to a mere examination of the various summarization techniques, but it examines the issues that arise in the use of these techniques taking into account the characteristics of the medical domain.

The structure of the survey is as follows. Section 2 presents a roadmap of summarization comprising the factors that have to be taken into account and the main techniques considered so far in the summarization literature. Section 3 presents different types of medical documents and the requirements that they introduce to the summarization process. Section 4 examines the techniques used so far for summarization in medical documents. Finally, Section 5 summarizes the most interesting remarks of this survey and presents promising paths for future research, while Section 6 concludes the paper.

## 2. Summarization roadmap

A summarization system in order to achieve its task takes into account several factors. These factors concern mainly the type of input documents, the purpose that the final summary should serve, and the possible ways of presenting a summary. Summary evaluation is also an important issue. These factors are examined in the following sub-sections. Various techniques that have been used so far for document summarization are also presented. This presentation is necessary for the examination of existing approaches to summarization from medical documents.

### *2.1. Summarization factors*

A detailed presentation of the factors that have to be taken into account for the development of a summarization system has been given in [7]. However, as it is noted there, all these factors are hard to define and therefore it is very difficult to capture them precisely enough to guide summarization in various applications. The following presentation of factors adopts the main categorization presented in [7]: input, purpose and output. For each of these categories, the factors considered as the most important ones are presented.

#### Input factors

The main factors in this category are the following.

**Single-document vs. multi-document:** This is the *unit input parameter* or the *span* parameter, as [7] and [8] respectively call it, which in simple words is the number of documents that the system has to summarize. In single-document summarization the system processes just one document at a time, whereas in multi-document summarization more than one document are processed by the system.

*Language:* Another input factor is the number of languages in which the input documents are written. So, a system can be *monolingual*, *multilingual* or *cross-lingual*. In the first case, the output language is the same as the input language. In the case of multilingual summarization systems, the output language is the same as the input language, but the system can handle a certain number of languages. In the final case of cross-lingual summarization, the system can accept a source text in a specific language and deliver the summary in another language, not necessarily the same as the input one.

*Text vs. multimedia summaries:* Another important factor is the medium used to represent the content of the input document(s), as well as the output summary. Thus, we have *text*, or *multimedia* (e.g. images, speech, video apart from textual content) summarization. The most studied case is, of course, text summarization. However, there are also summarization systems that deal, for example, with the summarization of broadcast news [9] and of diagrams [10].

### Purpose factors

These factors concern the possible uses of the summary, the potential readers of the summary, as well as the domain(s) that must be covered by the system.

*Informative vs. indicative summaries:* According to the function that the summary is supposed to serve when presented to its reader, it can either be *indicative* or *informative*. An indicative summary does not claim any role of substituting the source document(s). Its purpose is merely to alert its reader in relation to the contents of the original document(s), so that the reader can choose which of the original documents to read further. The purpose of an informative summary, on the other hand, is to substitute the original document(s) as far as coverage of information is concerned. Apart from the indicative and informative summaries, there are also *critical* summaries [7, 8], but, as far as we know, no actual summarization system creates critical summaries.

*Generic vs. user-oriented summaries:* This factor concerns the information a system needs to locate in order to produce a summary. Generic systems create a summary of a document or a set of documents taking into account all the information found in the documents. On the other hand, user-oriented systems try to create a summary of the information found in the document(s) which is relevant to a user query. In a sense, we can say that the query-oriented summarization systems are user-focused, adapting each time to the verbally expressed needs of the users, as viewed through the query they make or through their model (personalized summaries).

*General purpose vs. domain specific:* General purpose systems can be easily ported to a different domain (e.g. financial, medical). This can be done, for instance, by changing the resources that characterize the domain (e.g. key words, a domain-specific ontology), or by tuning specific parameters which concern the selection of the most appropriate techniques for the domain. On the other hand, domain-specific systems are able to process documents belonging to a specific domain.

### Output factors

These factors are related to the criteria that are used to judge the quality of the resulting summary as well as with the type of summary in terms of whether this is an extract from the original document(s) or an abstraction.

*Output quality:* The developer of a summarization system has to specify certain qualitative or quantitative criteria, which are related to the specific summarization task and the evaluation method that will be used (see section 2.2). Such criteria may be among others the completeness, the accuracy, the coherence of the resulting summary, etc. If accuracy is crucial for a specific task, then the system developer must tune its system accordingly, i.e. producing accurate summaries even though these do not contain all the relevant results.

*Extracts vs. abstracts:* Considering the relation that the summary has to the source document(s), a summary can either be an *extract* or an *abstract*. An extract involves the selection and verbatim inclusion of "material" from the source document(s) in the summary; this "material" is usually sentences, paragraphs or even phrases. The excerpted textual units can be included in the summary *verbatim*, or they can be processed further in order to smooth the text flow. An abstract, on the other hand, involves the identification of the most salient concepts prevalent in the source document(s), the fusion and the appropriate presentation of them, usually through Natural Language Generation.

## *2.2. Evaluation*

Although the summarization community considers evaluation as a critical issue, it still remains unclear *what* the evaluation criteria should be. This is mainly related to the subjective aspect of summarization, in terms of whether or not a summary is of ''good'' quality. Existing evaluation techniques can be split into two categories, *intrinsic* and *extrinsic ones* (see section 5 in [11]). An intrinsic method evaluates the outcome of a summarization system independently of the purpose that the summary is supposed to serve. An extrinsic evaluation, on the other hand, evaluates the produced summary in terms of a specific task.

An intrinsic method can measure the quality of the summary using criteria such as the integrity of its sentences, the existence of anaphors without their referents (for an extract), the summary readability (for an abstract), the fidelity of the summary compared to the source document(s). Another way to perform an intrinsic evaluation is to have human subjects create a "gold" summary, i.e. an ideal one, which will be compared to the summary created by the system. In this case evaluation can be more quantitative and measure things such as precision or recall. The problem with this approach is that it is usually difficult to make people agree on what constitutes a ''gold'' summary. One way to sidestep this problem is to employ a utility-based measure, in which a sentence is not assigned a Boolean value (belonging or not to the summary), but instead a value in a scale, according to the opinion of each judge. Those values are then averaged and the top sentences can be considered as forming the ``gold'' summary.

In an extrinsic method, the summary is evaluated in relation to the particular task it is supposed to serve. Thus, such an evaluation can greatly vary from system to system. Relevance assessment is an extrinsic method and is usually performed by showing judges a document (summary or source) and a topic and asking them to determine whether that document is relevant to the topic. If, on average, the choices for the source document and the summary are the same, then the summary scores high in the evaluation. Of course, instead of providing a single topic, a list of topics can be provided asking the judges to choose one of them. Another example is the evaluation of reading comprehension. In this case, judges are given a document (either the summary or the source document) and are asked a set of questions. Their answers to these questions determine their comprehension of the text. If the answers to the summary and the corresponding source document(s) are similar, then the summary is positively evaluated.

## *2.3. Summarization techniques*

Summarization techniques can be classified according to the factors presented in section 2.1. For example, they can be classified according to the number of input documents (single-document vs. multi-document), to the type of these documents (textual vs. multimedia), to the output types (extractive vs. abstractive), etc. In this section, the various summarization techniques are presented under the following classification:

- Extractive
- Abstractive
- Multidocument
- Multimedia

An extra category is added to include a technique that although it presents similarities with techniques in the other categories, it has the special characteristic of approaching summarization from a cognitive perspective aiming at simulating the human summarizers' tasks.

### 2.3.1. Extractive techniques

Two representative categories of extractive techniques implemented by existing systems (see Table 1) are presented below.

The first one concerns statistical techniques based on what [8: p 47–53] calls the Edmundsonian paradigm. In this paradigm, each sentence should be ranked in relation to the other sentences, so that the n highest ranked sentences could be extracted and form the summary. The ranking is normally based on a formula, which assigns a weight to each sentence based on various factors. For example, the cue phrases or keywords that the sentence contains, its location in the document, the fact that it may contain some non trivial words that are also found in the sections' headings of the document. The problem is that most of the times the produced summary is suffering from incoherencies (semantic gaps, anaphora problems). Some of the systems falling under this category post-process the produced summary using revision techniques in order to resolve such problems.

The second category of extractive techniques concerns the creation of a graph (or tree) representation of the document(s) to be summarized exploiting machine learning and/or language processing techniques. Several different representations can be used:

- The nodes of the graph are the paragraphs of the document to be summarized and the edges represent the *similarity* between the paragraphs they connect. The paragraphs corresponding to nodes with many edges can be extracted in order to form the summary of the document.
- The nodes of the graph are text items such as words, phrases, proper names and the edges are cohesion relationships between these items, such as *coreference, hypernymy*. Once the graph for the document is created, the most salient nodes in the graph can be located, based for example on a user query. The set of salient nodes can then be used to extract the corresponding sentences, paragraphs, or even sections that will form the summary.
- A tree representation can be created exploiting relations from the Rhetorical Structure Theory (RST) [12]. The tree nodes are sentences, which are connected using RST relations such as *elaboration, antithesis, etc*. Then in order to get the most salient sentences, the tree must be traversed in order to build a partial ordering of the sentences in terms of their importance. According to the target compression rate, the top *n* sentences can be extracted and presented as a summary.

Table 1 presents representative systems employing the extractive techniques presented above, taking into account the factors specified in Section 2.1. The *input* field concerns the number of input documents, their language, and whether they contain only text. The *purpose* field concerns whether the resulting summary is indicative or informative (in most of the cases it is difficult to judge this), generic or user-oriented, as well as whether the technique is a general purpose or a domain specific one. The *output* field concerns the "material" used to create the summary (sentences, paragraphs, sections). The table also includes a field about the specific *methods* used (e.g. statistics, language processing, use of revisions, etc.), as well as a field on the *evaluation* approach. The lack of the corresponding information in some field values denotes that a definite answer cannot be given.

**Table 1** Representative systems employing extractive techniques

|  | *Input* | *Purpose* | *Output* | *Method* | *Evaluation* |
| --- | --- | --- | --- | --- | --- |
| *[3]* | Single-document, English, Text | Generic, Domain-specific (technical papers) | Sentences | Statistics (Edmundsonian paradigm), No Revision |  |
| *[4]* | Single-document, English, Text | Generic, Domain-specific (scientific articles on specific topics) | Sentences | Statistics (Edmundsonian paradigm), Use of thematic keywords, No Revision | Intrinsic |
| *[13]* | Single-document, Multilingual, Text | Generic, Domain-specific (news) | Sentences | Statistics (Edmundsonian paradigm), No Revision | Intrinsic |
| *[14]* | Single-document, Multilingual, Text | User-oriented, Domain-specific (scientific and technical texts) | Sentences | Statistics (Edmundsonian paradigm), Use of thesauri, Revision |  |
| *[15]* | Multi-document, Multilingual (English, Chinese), Text | Generic, Domain-specific (news) | Sentences | Language processing (to identify keywords) | Extrinsic |
| *[16]* | Single-document, English, Text | Generic, General purpose | Paragraphs | Graph-based, Statistics (cosine similarity, vector space model) | Intrinsic |
| *[17]* | Multi-document, English, Text | User-oriented, General purpose | Text regions (sentences, paragraphs, sections) | Graph-based, cohesion relations, Language processing | Intrinsic, extrinsic |
| *[18, 19]* | Single-document, English, Text | Generic, Domain-specific (scientific articles) | Sentences | Tree-based, Language processing (to identify the RST relations markers) | Intrinsic, extrinsic |

### 2.3.2. Abstractive techniques

The most straightforward way of creating abstracts is to identify in a way the most important information in the document(s), appropriately encode it and then feed it to a Natural Language Generation (NLG) system [20, 21], which generates the summary. Two representative categories of abstractive techniques, implemented by existing systems (see Table 2), are presented below.

In the first category the process of identifying and encoding the most important information in the document(s) can be performed using prior knowledge about the structure of this information. This knowledge is represented through cognitive schemas such as frames, scripts, templates. Thus, in such cases, the summary produced is not a generic one, but a rather user-oriented one since the schema can be considered as a sort of user query. Different approaches in this category may be the following:

- Use of a *script*, i.e. a sort of a simple-structured template with slots identifying common important events over a domain. There is a separate script for each domain. When a document is processed, the corresponding script is activated and its slots filled with information from the document. The activation can be performed through the appearance of certain words, or by the activation of another script. After the script has been activated and filled, the summary can be generated using in most cases simple techniques (canned text generation vs. the more sophisticated NLG techniques) due to the rather simple structure of scripts.
- Use of a MUC[1]-like domain-specific template, a sort of a relational database, having a more complex structure compared to scripts. The template can be filled from a document using information extraction techniques. The filled templates can then be processed in order to transform them in an appropriate form for the NLG system. Processing is done using various semantic operators, such as *Change of Perspective, Contradiction, Addition/Elaboration, Refinement, Agreement,* etc.

The second category involves techniques that do not use prior knowledge about the structure of the important information to be used in the summary, but instead they produce a semantic representation of the document(s), which is then fed to the NLG system. Different approaches in this category may be the following:

- The documents are linguistically processed in order to identify noun phrases, verb phrases that can be linked to the concepts, attributes and relations of a domain-specific ontology. Ontology-based annotations can then be used to select the important document regions (sentences, paragraphs). These regions are then converted into some semantic representation using the results of the linguistic processing and the ontology-based annotations. This representation is then fed to a NLG system that produces the abstract.
- The summarization system can identify informational equivalent paragraphs in the input document(s) using clustering techniques. From each theme some representative sentences are extracted, which can be analyzed syntactically and then fed to a sentence generator in order to produce the abstract.

**Table 2** Representative systems employing abstractive techniques:

|  | *Input* | *Purpose* | *Output* | *Method* | *Evaluation* |
|---|---|---|---|---|---|
| [22] | Single-document, English, Text | Informative, User-oriented, Domain-specific | Scripts | Script activation, Canned Generation | |
| [23, 24] | Multi-document, English, Text | Informative, User-oriented, Domain-specific | Templates | Information Extraction, NLG | Evaluation of system components |
| [25] | Single-document, English, Text | Generic, Domain-specific (news articles) | Clusters | Syntactic processing of representative sentences, NLG | Intrinsic |
| [26] | Single-document, English, text | Informative, User-oriented, Domain-specific | Ontology-based representation | Syntactic processing of representative sentences, ontology-based annotation, NLG | Extrinsic |
| [27] | Single-document, Multilingual, Text | | Conceptual representation in UNL | Statistics (for scoring each UNL sentence), removing | |

---

[1] MUC (Message Understanding Conferences) were evaluation conferences for information extraction systems; see http://www.itl.nist.gov/iaui/894.02/related_projects/muc/index.html.

| | | | | redundant words, combining sentences | |
|---|---|---|---|---|---|

Table 2 presents representative systems employing abstractive techniques. Table 2 fields are the same with the fields in Table 1. There are differences between the two tables concerning the output and the method fields, which are filled with different values. The *output* field in the abstractive techniques concerns the "semantic representation" used to create the summary (scripts, templates, ontology-based representation, clusters of informational equivalent document regions). The *method* field is filled with the specific methods used (script activation, information extraction, syntactic processing, NLG, etc.).

### 2.3.3. Multi-document summarization techniques

[2] define multi-document summarization as the process of producing a single summary of a set of related source documents. As they note, this is a relatively new field where three major problems are introduced: (1) recognizing and coping with redundancy, (2) identifying important differences among documents, and (3) ensuring summary coherence.

While a technique extracting textual units – such as sentences – from a single document, may cope with redundancy and preserve the coherence of the original document, extracting textual units from multiple documents increases the redundancies and incoherencies, since textual units are not previously connected across documents. Thus, abstractive techniques seem more appropriate for multi-document summarization. However, extractive techniques can also be used followed by a post-processing stage in order to ensure summary coherence and cope with redundancy. In both cases, the first step is to identify those documents talking about a specific topic. This can be done, using classification techniques in terms of relevance to a query or existing topic models, or using clustering techniques. The next step is to identify the important information to be added in the summary in the group of the topic-specific documents.

In the extractive-based category, a system can apply firstly extractive techniques to each document separately in order to locate and rank the most important document regions (phrases, sentences, paragraphs). The highest ranked regions from all the documents can then be combined and re-ranked using similarity measures, in order to minimize redundancy. Some cohesion rules can then be applied to the final set of document regions in order to produce the summary. Another approach would be to work with all the documents from the beginning. For instance, the topic model produced from the group of input documents can be compared to sentence vectors from all the documents in order to get the, most similar to that topic, sentences.

The abstractive-based category can also involve extractive techniques at a pre-processing stage. In such a case, the extracted document regions are linguistically processed in order to be converted into some representation, which can be used by a NLG system, which will then produce the abstract. Another approach involves the establishment and use of a set of intra- and inter-document relationships, which could hold between various textual units of the documents, such as words, sentences, paragraphs or whole documents, and which could guide the identification of the most salient information across the multiple documents. Such relationships concern not only the similarities across documents, but also their differences (e.g. e*quivalence, contradiction, elaboration,* etc.).

To cope with the inherent problems of multi-document summarization (redundancy, incoherencies), a different output representation can be used instead of producing a summary containing, for instance, the most salient sentences across documents or even their abstraction. [28] use a scatter plot per topic presenting the extracted sentences visually to the user.

Table 3 includes representative examples of multi-document summarization techniques. Compared to Tables 1 and 2, there are differences concerning the values in the output and the method fields. The *output* field concerns whether the output is an extract or abstract or even another representation format. The method field is filled with the specific methods used (statistical, language processing, revision stage, use of intra- and inter document relationships).

**Table 3** Representative systems employing multi-document summarization techniques

| | *Input* | *Purpose* | *Output* | *Method* | *Evaluation* |
|---|---|---|---|---|---|
| *[29]* | Multi-document, English, Text | User-oriented, General purpose | Extracts | Statistical (Multi-Document Maximal Marginal Relevance –MD-MRD), Revision | Intrinsic |
| *[30]* | Multi-document, | Generic, General | Extracts | Statistical (Support Vector | Intrinsic |

| | English, Text | purpose | | Machine, MD-MRD) | |
|---|---|---|---|---|---|
| [28] | Multi-document, English, Text | Generic, General purpose | A scatter plot of the most salient sentences | Statistics (vector space model), language processing | |
| [31] | Multi-document, English, Text | Generic, General purpose | | Intra- and inter-document relationships | |

### 2.3.4. Multimedia summarization techniques

In this section, work on multimedia summarization is discussed. This involves such diverse fields as dialog summarization, summarization of diagrams, and video summarization. Due to the limited number of relevant works and their field-specific features, only one approach per field is presented.

*Dialogue summarization:* [32, 33] works on dialogue summarization in unrestricted domains and genres. Zechner's works with transcripts of human dialogs, which are generated either manually or automatically (in this case Zechner also addresses the issue of speech recognition errors). The major steps in Zechner's work are the following. In the first step, *input tokenization*, all noise is removed and the transcript is tokenized. *Disfluency detection* follows in which false starts, restarts or repairs and filled pauses are annotated and corrected. In this step the boundaries of the speakers' sentences are also detected. In the next step, *cross-speaker information linking*, pairs of question-answer among the speakers are identified and annotated. Those pairs are extracted together later when producing the summary, thus making the resulting summary more coherent and informative. The following step is *topic segmentation* where segments of the discussion on a certain topic are identified and a list of keywords, for each topic, is extracted. The final step, *sentence ranking and selection*, uses a version of the Maximal Marginal Relevance (MMR) algorithm [34] to create a summary of extracted sentences, for each topic.

*Diagram summarization:* [10] presented a preliminary and only partially implemented work on diagram summarization, yet quite innovative. His aim was to present a summary of the diagrams in a scientific paper, either by *selecting* (*i.e.* extracting) one or more diagrams from the paper or by *distilling* (*i.e.* simplifying) a diagram, or even by *merging* several diagrams. Futrelle assumes that a structural description of the diagrams can be obtained as metadata from the author or by parsing the diagrams. In order to achieve his goals he takes into account not only the structural description of each diagram, but the text in its caption or in the diagram itself.

*Video summarization:* [9] extract information from various media in order to provide a summary of a broadcast news story. They use MITRE's Broadcast News Navigator (BNN; [35]) which helps searching and browsing of news stories, not from just one channel, but from a variety of sources. They use silence and speaker change from the audio, anchor and logo detection from the video and closed-captioned text in order to segment the stream into news stories. For the purpose of presenting a summary, they experiment with a diversity of presentation methods, which include mixed-media. The first one is key-frames, *i.e.* important single frames and shots *i.e.* an important sequence of frames. The second one is extracted single sentences, based on weights according to the presence of named entities (organizations, persons and places). The final one is named entities keywords.

### 2.3.5. Summarization from a cognitive science perspective

The techniques presented so far pay little attention to the ways used by humans to create the summaries themselves. [36, 37, 38, 39] on the other hand, try to simulate the human cognitive process of professional summarizers. They aim at developing an empirical model for summarization, based on professional summarizers, and to implement this model into a system, which will "imitate" human process of summarizing. For this purpose, they recruited six professional summarizers who worked on 9 summarization processes. The whole process included the division of the tasks of the professional summarizers into sub-tasks, the interpretation of each sub-task into a more formal framework (i.e. giving a name, functional definition, etc.), and the hierarchical organization of the resulting strategies according to their function. Despite the diversity of the technical background and cognitive profile of each professional summarizer and their idiosyncrasies on creating summaries, the results are quite stunning. Quoting from [36: p 129]: "*83 strategies are used by all experts of the group, 60 strategies are shared by five experts, another 62 strategies are common knowledge of four summarizing experts, 79 strategies belong to the repertory of three summariza-*

*tion experts, 101 strategies are used by two experts, and 167 strategies are individual.*" From all these strategies, 79 agents, which simulate them, were finally implemented in the SimSum system [36]. SimSum is implemented as an object-oriented blackboard, which involves 79 object-oriented agents, each one performing a relatively simple task. For instance, the *Context agent* checks whether the context conditions of the query are met by an input document, the *TexttoProposition agent* transforms input sentences into propositions known to the domain ontology, the *Redundancy agent* checks if a proposition has already been introduced. All the agents cooperate in order to deliver the summary. They can access a common knowledge database, which contains the text and ontology concepts. Furthermore several RST relations have been implemented for discourse level structures of the text.

## 3. The medical domain

Medical Informatics represents the core theories, concepts and techniques of information applications in medicine. It involves four different levels depending on the focus from the cell to the population [40]:

- Bioinformatics concerns molecular and cellular processes, such as gene sequences;
- Imaging informatics concerns tissues and organs, such as radiology imaging systems;
- Clinical informatics concerns clinicians and patients, involving applications of various clinical specialties;
- Public health informatics concerns populations involving applications such as the disease surveillance systems

Medical information distributed through all the above levels concerns various document types: scientific articles, electronic medical records, semi-structured databases, web documents, e-mailed reports, X-rays images, videos. The characteristics of each document type have to be taken into account in the development of a summarization system. Scientific articles are mainly composed of text and, in several cases, they have a sectioning that can be exploited by a summarization system. Electronic medical records contain structured data, apart from free text. Web documents may appear in health directories and catalogs, which need to be searched first in order to locate the interesting web pages. The web pages layout is also another factor that needs to be taken into account. E-mailed reports are mainly free text without any other structure. X-rays images, videos such as echocardiograms, represent a completely different document type, where text may not be included at all or may be a part of an image.

Compared to other domains, medical documents have certain unique characteristics that make very challenging and attractive their analysis. Uniqueness of medical documents is due to their volume, their heterogeneity, as well as due to the fact that they are the most rewarding documents to analyze, especially those concerning human medical information due to the expected social benefits (see [41]).

*Scientific articles*

The number of scientific journals in the fields of health and biomedicine is unmanageably large, even for a single specialty, making very difficult for physicians and researchers to stay informed of the new results reported in their fields. Scientific articles may contain apart from text, structured data (e.g. tables), graphs, or images. Therefore, depending on the summarization task, the system may have to process various types of data. It may be the case, for instance, that important information (e.g. experimental results) is found in a table and needs to be located and added to the summary. The article layout could also be exploited, since a large number of articles reporting on experimental results have an almost standard sectioning, the order of the sections being I*ntroduction, Methods, Statistical Analysis, Results, Discussion, Previous Work, Limitations of the Study,* and *Conclusions*. A study of the types of scientific articles on the various fields of medicine is necessary for their processing either for summarization or for other language processing tasks.

*Databases of abstracts*

Despite the plethora of medical journals, most of them are not freely accessible over the Web, due to copyright reasons. Luckily, there are other online databases, which contain the abstracts and citation information of most articles on the general field of Medicine. One such online database is MEDLINE[2], which contains abstracts from more than 3500 journals. MEDLINE provides key-word searches and returns abstracts that contain the keywords. The abstracts are indexed according to the Medical Subject Headings (MeSH)[3] thesaurus. Apart from the access to the abstracts, MEDLINE also provides full citations of the articles along with links to the articles themselves, in case they are online. [40] have created a corpus from Medline, which consists of titles and abstracts from articles of 270 Medical Journals over a period of five years. This corpus is annotated with information such as the Medline identifier, MeSH terms assigned by humans, title, abstract, publication type, source and authors. The corpus was created for the experiments described in [40]. Abstracts represent a different type of document, since they are only composed of text and metadata (such as the MeSH annotations in MEDLINE). A summarization system must be able to exploit these metadata in several ways. For instance, in multi-document summarization, the system must be able to locate first those abstracts that discuss a specific topic, and topic information is found on the metadata of the abstracts. On the other hand, a sublanguage analysis of the abstracts may be necessary in order to identify certain characteristics that may affect the performance of the language processing application. As it is noted in Ariadne Genomics NLP white paper[4], their sub-language analysis indicated that "MEDLINE sentences often include idiosyncratic linguistic constructs not necessarily reflected in generalized English grammar". This explains, as they claim, why existing syntactic parsers with general grammar are not suitable for dealing with this type of text.

---

[2] http://www.ncbi.nlm.nih.gov/entrez/query.fcgi

*Semi-structured databases*

A number of databases have been built to provide access to biomedical information, such as information about protein function and cellular pathways. More than 280 semi-structured databases exist currently[5]. Some examples are the following:
- FlyBase: fruit fly genes and proteins
- Mouse Genome Database (MGB)
- Protein Information Resource (PIR)
- DIP: The database of interacting proteins[6]

Let us take DIP, for example, which provides an integrated set of tools for browsing and extracting information about protein interaction networks. As it is reported in [42], DIP database is implemented as a relational database composed of four tables: protein table, interaction table, method table, reference table. The reference table lists all the references to different articles that demonstrate protein interactions and link them to MEDLINE database. Therefore, a summarization system that exploits information from DIP should be able to employ DIP tools for browsing the database and even access relevant MEDLINE abstracts (see previous discussion) through the reference table.

*Web documents*

During the last few years, a number of specialized health directories and catalogs (portals) have been created, such as CliniWeb[7], HON[8], CISMeF[9], Medical Matrix, Yahoo Health, HealthFinder. Some of these catalogs, including the first three of the above, are additionally indexed with the MeSH Thesaurus. This allows complex queries to be stated, which exploit the hierarchical structure of the MeSH. CliniWeb, for instance, provides clinically oriented information, including:
1. Cataloging content per-page basis;
2. Including only pages that have clinical content, i.e., excluding individual and institutional home pages, advertisements, and lists of links;
3. Indexing with a higher level of specificity, using MeSH as opposed to broad subject categories such as Orthopedics or Cancer. CliniWeb provides access to Web pages manually indexed by a large subset (trees A-G) of MeSH, including the major trees, Diseases, Anatomy, and Chemicals and Drugs.

---

[3] http://www.nlm.nih.gov/mesh/meshhome.html
[4] http://host.ariadnegenomics.com/downloads/
[5] See KDD Cup 2002 Task1:Information Extraction from Biomedical Articles (http://www.biostat.wisc.edu/~craven/kddcup/)
[6] http://dip.doe-mbi.ucla.edu
[7] http://www.ohsu.edu/cliniweb/
[8] http://www.chu-rouen.fr/cismef/
[9] http://www.chu-rouen.fr/cismef

Another example of a health resources portal is CISMeF, which aims to describe and index the main French- language health resources to assist health professionals and consumers in their search for electronic information available on the Internet. In April 2002, the number of indexed resources totaled over 9,600 with a mean of 50 new sites each week. CISMeF uses two standard tools for organizing information: the Medline bibliographic database MeSH thesaurus and several metadata element sets, including the Dublin Core. To index resources, CISMeF uses four different concepts: "meta-term", keyword, subheading, and resource type. CISMeF contains a thematic index, including medical specialities and an alphabetic index.

Web pages included in catalogs, such as the above ones, form a different type of medical documents. These are .html pages, which may contain information from other media apart from text (e.g. images, videos). Even in the case that they contain text only, there will be links pointing to other relevant pages with interesting information for the summarization task, there may be interesting information stored in a table, etc. Web page layout should also be taken into account in order to locate interesting information inside the web page, especially in the case that the catalog pages are generated dynamically from a database. In addition, the identification of web pages that are relevant to the summarization task, demands the use of web spidering techniques. Therefore, as it is the case of web information retrieval and extraction tasks in other domains (see the results of the CROSSMARC project[10] in [43]), summarizing from web documents needs to take into account the web catalog and web pages structure and features. It must also be able to exploit metadata information (e.g. MeSH annotations) already used by some of the existing web catalogs. However, even in the cases where a web catalog is not indexed, the summarization system must be able to employ existing medical ontologies, thesauri or lexica. This is essential for a scientific domain with rich terminology, where the information to be extracted or summarized needs to be as precise as possible.

*E-mailed reports*

With the advent of the Web, several web-based services, whose purpose is the exchange of opinions and news, have emerged. For example, ProMED-mail[11] is a public free service, which promotes the exchange of news concerning bursts of epidemics. Other non-free services, such as MDLinx[12], provide physicians and researchers with the opportunity to subscribe and receive alerts concerning new findings on their specialty fields, described in journal articles. The use of e-mailed reports for a fast dissemination of epidemiological information by the Internet shows an increasing success for monitoring epidemiological events [44]. The descriptive possibilities of these reports and their ability to deal with unattended situations make them competitive for reporting emerging infectious disease outbreaks and unusual disease patterns, including biological threats. However, as [45] note, analysts cannot feasibly acquire, manage, and digest the vast amount of information available through emailed reports or other information sources 24 hours a day, seven days a week. In addition, access to foreign language documents as well as the local news of other countries is generally limited. Even when foreign language news is available, it is usually no longer current by the time it gets translated and reaches the hands of an analyst. This very real problem raises an urgent need for the development of automated support for global tracking of infectious disease outbreaks and emerging biological threats.

ProMED-mail is a service monitoring news on infection disease outbreaks around the world, seven days a week. By providing early warning of outbreaks of emerging and re-emerging diseases, ProMED aims at enabling public health precautions at all levels in a timely manner to prevent epidemic transmission and to save lives. ProMED sources of information include –among others- media reports, official reports, online summaries, local observers. Reports are also contributed by ProMED-mail subscribers. A team of expert moderators investigate reports before posting them to the network. Reports are distributed by e-mail to subscribers and posted on the ProMED-mail web site. ProMED-mail currently reaches over 30,000 subscribers in 150 countries. ProMED-mail is also available, apart from English, in Portuguese and in Spanish. Both of these lists cover disease news and topics relevant to Portuguese- and Spanish-speaking countries, respectively.

E-mailed reports for monitoring infectious disease outbreaks and emerging biological threats represent a different type of medical documents. Such reports may contain apart from raw text, various types of information in attached files. The fact that these reports may be in several languages or may point to other

---

[10] http://www.iit.demokritos.gr/skel/crossmarc
[11] http://www.promedmail.org
[12] http://www.mdlinx.com/

sources, such as local news, makes the summarization task even more difficult. A sublanguage analysis may also be necessary for these types of documents since they often follow a specific writing style and structure. Medical terminology should also be taken into account, as it is the case in the other document types, exploiting existing medical resources for the specific diseases or biological threats.

*Electronic medical records*

Most hospitals keep a record for each of their patients. Usually the records contain data of patients in a standard structured form, with predefined fields or tabular representations, as well as free-text fields containing unstructured information, usually doctors' reports about their patients (either written reports or the result of dictation). As [46] note, a patient record for any single patient consists of many individual reports, collected during a visit to hospital. For some patients, this can be up to several hundred reports.

A system summarizing information from medical records needs to take several factors into account. It must be able to process the free text reports, which may be problematic due to the specific sub-language used by the clinicians or due to the fact that the report is the result of dictation. Information from written reports may also be combined with information existing in structured data (tables, graphs) or even information existing in other media (e.g. X-rays images, videos). The situation, however, may become even more complex for a summarization system that aims to summarize information for a clinician, which is collected not only from the patient record but also from other records (similar cases to the specific patient), from relevant scientific articles or abstracts from journals or databases respectively. If a summarization system is to be integrated into the busy clinical workflow, it must provide the clinician with such facilities.

*Multimedia documents*

Apart from documents in textual form, physicians and researchers produce and use several other documents, which are multimedia in nature. Such documents can be *graphs*, such as cardiograms, *images* such as X-Rays, etc., *videos*, such as the various echograms, *e.g.* echocardiograms, echoencephalograms, etc., or the medical videos used mainly for educational purposes, e.g. videos of clinical operations or videos of dialogs between the doctor and the patient. Most of these documents are now transcribed and stored in digital form, even connected to the specific patient record, giving the users the ability to search and access them much faster than in the past. This is a completely different type of medical documents, which contain very interesting information that must be added to a summary. Techniques from areas other than language processing, such as image processing and video analysis must also be employed in order to locate the information to be included in the summary. In addition, several of these multimedia documents are also linked with free text reports, which must also be used by the summarization system. As noted in the discussion on electronic medical records, a summarization system integrated in the clinical workflow must be able to handle such documents. Concluding, in the medical domain, processing of multimedia documents is crucial for summarization and in general for information retrieval and extraction applications.

## 4. Summarization techniques in the medical domain

Most of the researchers extend to the medical domain the techniques already used in other domains. Based on the categorization given earlier in section 2.3, the techniques used in the medical domain are classified under the following categories:

- ➢ Extractive single-document summarization
- ➢ Abstractive single-document summarization
- ➢ Extractive multi-document summarization
- ➢ Abstractive multi-document summarization
- ➢ Multimedia summarization
- ➢ Cognitive model based summarization

In the following sub-sections, various summarization projects/systems are presented based on this categorization.

## 4.1. Extractive single-document summarization

One of the projects belonging in this category is MiTAP [45]. The aim of MiTAP (MITRE Text and Audio Processing)[13] is to monitor infectious disease outbreaks or other biological threats by monitoring multiple information sources such as epidemiological reports, newswire feeds, email, online news, television news and radio news in multiple languages. All the captured information is filtered and the resulting information is normalized. Each normalized article is passed through a zoner that uses human-created rules to identify the source, date, and other fields such as the article title and body. The zoned messages are processed to identify paragraph, sentence and word boundaries as well as part of-speech tags. The processed messages are then fed into a named entity recognizer, which identifies person, organization and location names as well as dates, diseases, and victim descriptions using human-created rules. Finally, the document is processed by WebSumm [47], which generates a summary out of modified versions of extracted sentences. For non-English sources, a machine translation system is used to translate the messages automatically into English. In addition to single-document summarization, MiTAP has recently incorporated two types of multi-document summarization: Newsblaster[14] [46] automatically clusters articles and generates summaries based on each cluster, Alias I[15] produces summaries on particular entities and generates daily top ten lists of diseases in the news.

Another project is MUSI [49]. MUSI stands for "MUltilingual Summarization for the Internet" and it is a *cross-lingual* summarization system, which uses articles from *The Journal of Anaesthesiology* as input. The journal is freely accessible online[16] and its articles are written in Italian and English. MUSI takes those articles and creates summaries from them in French and German. The system is query-based and it extracts sentences from the input article according to the following criteria: *cue phrases, position of the sentences, query words* and *compression rate*. That is, MUSI follows the Edmundsonian paradigm for the selection of the sentences. Once the sentences have been extracted, two approaches can be followed: either they are used as they are to form the extractive summary or they are converted into a semantic representation to produce an abstractive summary.

A third project exploiting extractive techniques is presented in (Johnson et al. 2002). The most important aspect of this approach is that it ranks the extracted sentences according to the so-called cluster signature of the document. More specifically, their prototype system takes medical documents (result of a query using a search engine) as input and clusters them into groups. These groups are then analyzed for features with high support, called key features, forming a cluster signature that best characterizes each document group. The summary is generated by matching the cluster signature to each sentence of the document to be summarized. Both the sentence and the cluster signature are represented using a vector space model. The ranked sentences are then selected and presented to the user as a summary. [50] used for their experiments abstracts and full texts from the *Journal of the American Medical Association*.

## 4.2. Abstractive single-document summarization

MUSI [49] is a system generating either extractive summaries (see the previous section) or abstractive ones. In the case of abstractive summarization, after the system has selected the sentences, it converts them into a predicate-argument structure representation, instead of simply presenting them to the user. The steps in achieving that representation are: tokenization, morphological analysis, shallow syntactic parsing, chunking, dependency analysis and mapping to the internal representation. After the representation has been achieved, they create the summaries of those extracted sentences using the natural language generation (NLG) system Lexigen [51] for the French language and TG/2 [52] for German. The generation systems produce indicative summaries of the document content. Summaries include both translated portions of the extracted sentences, and "meta-statements" about the original document. The latter provide the user with additional optional information about the content and structure of the source text, the relevance of the extracted pieces of information as well as of the whole document with respect to the query, etc. Users can customize the summary length, as well as some other aspects concerning style and presentation.

---

[13] For more information on MiTAP, visit http://tides2000.mitre.org
[14] http://www.cs.columbia.edu/nlp/newsblaster
[15] http://www.alias-i.com
[16] You can access this online journal at http://anestit.unipa.it/esiait/esiaing/esianuming.htm

TRESTLE (Text Retrieval Extraction and Summarisation Technologies for Large Enterprises) is a system, which produces single sentence summaries of Scrip[17] pharmaceutical newsletters [53]. Their system is in essence an Information Extraction system, which relies heavily on Named Entity (NE) recognition. For this system, also drug names and diseases are named entities, apart from the classical ones, such as organization, person and location. TRESTLE allows users to navigate through the Scrip articles, and thus find the information that they are interested in, using the named entities that the system has extracted, which are links to the original articles from which the NEs have been extracted. Apart from this, TRESTLE also creates single sentence summaries for each newsletter from the *template* that was filled by the Information Extraction process. A link is also provided to the original newsletter.

### *4.3. Extractive multi-document summarization*

Although the production of summaries from multiple documents is usually done with abstractive techniques, [54, 55] follow a different approach. They argue that different types of summaries, such as indicative or informative, serve different informational purposes and both can be useful, and that extracting sentences for the creation of an informative multi-document summary "*is well accepted since it is simple, fast and easy to evaluate*". Their system, Centrifuser, which is the summarization engine of the PERSIVAL (PErsonilized Retrieval and Summarization of Image, Video and Language) project[18], produces both indicative and informative multi-document summaries, with the aim of highlighting the similarities and differences among the documents.

The input to the Centrifuser is articles, retrieved by the search engine of the PERSIVAL system according to the patient record and the user query. For each article they create a *topic tree* which depicts the sectioning of each article. A *composite topic tree* is then created by merging together all the topic trees and adding details to each node such as relative typicality (*i.e.* how typical is that topic compared to the rest of the topics), position within the article, and various lexical forms in which it may be expressed [54]. In the next step they try to match the nodes of the topic trees with the query. The matched nodes do not contain any text, but instead they point to sections in the original documents, from which the most representative sentences should be extracted. Since the compression rate posed will not always allow for each topic to receive a sentence, the first step is to choose which topics are going to receive a sentence. In the next step they choose for each topic the representative sentences. The final step for the creation of the summary involves the ordering of those sentences, which is achieved by first ordering the topics according to each topic's typicality, and then ordering the sentences themselves inside every topic, according to the physical position of every sentence.

### *4.4. Abstractive multi-document summarization*

Apart from informative extractive multi-document summaries, Centrifuser creates indicative abstractive multi-document summaries, as well, which are used by the PERSIVAL users for searching papers. As noted in section 4.3, the approach of [54, 55] leads to nodes in the topic trees which match with the query of the user. This, they argue, can be the first phase for Natural Language Generation (NLG). In the next step of NLG, which they call planning, they try to figure out which nodes of the topic trees they will summarize. To achieve this, they determine which nodes are *relevant, irrelevant* and *intricate*, based on how deep the nodes are, compared with the query node, i.e. the node that matches the user query. Thus, nodes that are descendants of the query node and are below depth k are considered intricate, above depth k relevant and all the other nodes (i.e. the ones that are not descendants of the query node) are irrelevant. In the final NLG step, *realization*, the ordered information is converted to text. For a more thorough treatment of Centrifuser, see [56].

Apart from the abstractive indicative summaries that Centrifuser produces, PERSIVAL produces another type of abstractive summary [1, 57]. These summaries are not concerned with highlighting the similarities and differences among several medical articles, but with the creation of an informative abstractive summary. That summary is tailored according to the preferences of two different types of users: the physician, the patient or her relatives.

---

[17] See www.pjpub.co.uk for more information on *Scrip* newsletters.
[18] See http://persival.cs.columbia.edu/

The system should identify in the documents and extract tuples of the form: *(Parameter(s), Finding, Relation)*. The relations can be any of the following six types: *association, prediction, risk, absence of association, absence of prediction and absence of risk*. They call those tuples *results*. [57] using empirical methods, *i.e.* interviews with the physicians, concluded that a summary should fulfill the following qualitative criteria, in relation to the results:

- *Completeness and accuracy.* The results should be *complete* and *accurate*, in the sense that all the relevant results and only them should be included.

- *Repetitions and contradictions.* The system should identify repetitions and contradictions among the results. In order to do so, [57] have created a representation of the results, which allows them to identify relations such as subsumption and contradiction among the results.

- *Coherence and cohesion.* Coherence for [57] is established by "accurate aggregation and ordering of the related results". Cohesion is defined as follows: "two sentences are part of the same paragraph, if and only if they are related." Related are the sentences that present either the same finding, or the same parameter(s).

The system described in [57] takes input from three different sources:

- *Patient record.* In general, the patient record consists of structured documents, usually in tabular form, and unstructured documents, and sometimes it can be very large.

- *Journal medical articles.* Their system takes as input a vast amount of online articles from medical journals on the field of cardiology. In fact, the articles that are input to the system are the ones that *globally match* the patient, *i.e.* the ones that contain information relevant to the patient.

- *The user query.* Although the physician's query is posed in natural language, the system does not try to fully understand the question and give an answer, but instead gives as much information as possible about the question using some of the query keywords.

The input articles are first classified automatically into three categories: *prognosis, treatment, diagnosis.* The next step involves the identification and extraction of the *results*, *i.e.* the tuples mentioned above. For this purpose, the authors are exploiting the "rigid", as they call it, structure of the medical articles. This means that they try to locate the *Results* section and *select* the sentences that are relevant to the patient. The selected sentences are then passed to the extraction module, that extracts in a template form, the following information: the finding(s), the parameters, the relation, the degree of dependence of the parameters, the article and the sentence it has been extracted from and various other minor information. The templates are filled with the aid of handcrafted patterns.

The next step involves the determination of which portions, if any, of the extracted parameters are relevant with the patient record. After that, the resulting templates are merged and ordered. To achieve this, the templates are rendered into an internal "semantic" representation, in the form of a graph. From this graph, they are able to identify repetitions and contradictions. A **repetition** occurs if two nodes are connected by more than one vertex, and the vertices have "similar" types. What is similar and what is not has been "established" in interviews with physicians. A **contradiction** occurs in the same situation, but now the vertices have different types. Repetitions and contradictions are used in order to create a more coherent summary. With this method they manage to perform the merging of the templates. For the ordering, they use the following criteria:
- Query based: a relation that answers the user query is weighted higher.
- Salience based: recitations and contradictions are weighted higher.
- Domain based: studies with physicians show that some relation types are more interesting than others. For instance a risk relation is weighted higher than an association relation.
- Source based: dependent relations from the same template are presented together.

The final step involves the creation of the summary, through NLG techniques. In the final summary, all the medical terms are hyperlinked to their definitions. This is achieved by connecting the system of [57] with DEFINDER, a text mining tool for extracting definitions of terms from medical articles (see [58]).

*4.5. Multimedia summarization*

[59] and [60] present systems performing summarization of documents which have multimedia content, echocardiograms and medical videos respectively.

The work presented in [50] is part of the PERSIVAL project mentioned above. In their study they are concerned with echocardiograms (ECGs). ECGs are usually videotaped for archival purposes and recently they have started to be transcribed into a digital format, which helps clinicians, and facilitates the task of summarizing them. Summarizing an ECG, and video in general as seen in the work of [9], involves extracting the most interesting video frames, which are called *key frames*, which enable the user to easily navigate through the ECGs and view their essential parts. For [59] summarizing an ECG involves two things: parsing the ECG and selecting the key frames. The aim of the *parser* is to temporally segment the sequences of the video into smaller units, which are called *shots*. A shot is a sequence of frames in which the camera is uninterrupted. In the context of ECG videos, a shot corresponds to a single position and angle of the ultrasound transducer. The method they use for the parsing is a special case of the algorithm presented in [61]. The next step is the *key-frame selection,* which extracts the most informative (important) frames in the sequence of the video. After mentioning several methods for extracting key-frames, they conclude that for the context of ECGs key-frames are "the local extrema of the cardiac periodic expansive-contractive motion", since "the time at which the cardiac motion changes from expansive to contractive corresponds to the end-diastole and the time at which the motion changes from contractive to expansive corresponds to end-systole". Having performed the above two tasks, they create two summaries which they call *static* and *dynamic*.

- *Static summary.* This summary, in essence, is constituted from the selection of the extracted key-frames, and it is useful for browsing the content of the echo video.

- *Dynamic summary.* This summary, also called clinical summary among the clinicians, is a concatenation of the small extracted sequences of the video. They chose to extract one (or more, based on the needs of the clinicians) cycle of the heart motion, known also as R-R cycle. By joining those segments of videos they create the dynamic summary.

[60], follow a similar approach to [59] in order to parse the video stream into physical units. Then video group detection, scene detection and clustering strategies are used to mine the video content structure. Various visual and audio feature processing techniques are utilized to detect some semantic cues, such as slides, face and speaker changes, etc. within the video, and these detection results are joined together to mine three types of events from the detected video scenes (presentations by doctors or experts on video topics, clinical operations to present details of diseases, and dialogs between doctors and patients). Based on mined video content structure and event information, a scalable video skimming and summarization tool, ClassMiner, has been constructed to visualize the video overview and help users access video content. Their system utilizes a four layer video skimming, where levels 4 through 1 consist of representative shots of clustered scenes, all scenes, all groups, and all shots of the video, respectively.

*4.6. Cognitive model based summarization*

Based on the cognitive model used in the SimSum system (see sub-section 2.3.5), [62, 63] presented its extension, the SummIt-BMT, which is concerned with the summarization of MEDLINE abstracts and articles for bone marrow transplantation, a specialized field of internal medicine. SummIt-BMT is a query based summarization system. In general, the summarization process is the following:
1. A user forms a search scenario using concepts from the domain ontology.
2. This scenario is mapped to a MEDLINE query. If the outcome of the query points to journal articles, they are included in the results.
3. A text retrieval component identifies the interesting pieces of text in the results.
4. Those pieces are summarized in relation to the query scenario. Links to the original articles are also given.

Although SummIt-BMT is based on SimSum it differs from it in several ways. It is not a presentational model anymore but a functional one. Thus, agents simulating lower level cognitive processes have been replaced by functional ones. Text production agents have been removed since SummIt-BMT does not produce smooth text, but organized text clips that are linked to their source positions. As the application field is bone marrow transplantation (BMT), a BMT ontology was set up. Although several medical on-tologies existed which are loosely related to the BMT field, a BMT-specific ontology was created due to the fact that the existing ontologies did not contain enough deep BMT knowledge for text knowledge processing. The ontology they created is very important for Summit-BMT, since it is being used in almost all the stages of the summarization process.

A scenario interface reflecting everyday situations of BMT physicians [64] helps users to state their queries. Users fill in ontology concepts, which are for their convenience equipped with definitions and explanations assembled from various sources on the web. From scenario forms and user-selected ontology terms the system obtains structured queries in the predicate-logic form that it can "understand". Queries are given to the search engines, which return a set of documents, abstracts and maybe journal articles, from MEDLINE. The retrieved documents are checked for possible relevance by a text passage retrieval component. Irrelevant documents are discarded. From the final set of documents, the summarization agents take the positive passages from text passage retrieval, represent their phrases and sentences in a predicate logic form, and examine them with human-style criteria: whether they are related to the user query, whether they are redundant, and so on. The agents remove items that do not meet their relevance criteria.

The following Table summarizes the main features of the projects/systems presented in section 4.

**Table 4** Summarization systems from medical documents

|  | *Input* | *Purpose* | *Output* | *Method* | *Evaluation* |
|---|---|---|---|---|---|
| *[45]* | Single-document (also multi-document), Mutilingual, Multimedia (text, audio, video) | Indicative, User-oriented, Domain-specific | Sentences (extracts) | Language processing (named entity recognition, machine translation), Machine learning | Extrinsic |
| *[49]* | Single-document, Multilingual, Text | Indicative, User-oriented, Domain-specific | Sentences (extracts), abstracts | Statistics (sentences extraction), Language processing (semantic representation for abstraction) | Intrinsic, extrinsic |
| *[50]* | Single-document, Monolingual, Text | Indicative, Generic, Domain-specific | Sentences (extracts) | Statistics (vector space model) |  |
| *[53]* | Single-document, Monolingual, Text | Indicative, Generic, Domain-specific | Abstracts | Language processing (information extraction) |  |
| *[56]* | Multi-document, Monolingual, Text | Indicative-Informative, Generic, Domain-specific | Extracts, Abstracts | Statistics (clustering using similarity measures), Language processing | Extrinsic |
| *[57]* | Multi-document, Monolingual, Text | Informative, User-oriented, Domain-specific | Abstracts | Language processing (information extraction, NLG) |  |
| *[59]* | Single-document, Video (echocardiograms) | Generic, Domain-specific | Video sequences (extracts) | Image and video processing |  |
| *[60]* | Single-document, Video (clinical operations, dialogues, presentations) | Generic, Domain-specific | Video sequences (extracts) | Image and video processing |  |
| *[62, 63]* | Multi-document, Monolingual, Text | Informative, User-oriented, Domain-specific | Abstracts | Agents simulating summarization tasks, Language processing |  |

## 5. Promising paths for future research

Although initial work on Summarization dates back to the late 50's and 60's (*e.g.* [3, 4]), most research on the field has been performed during the last few years. The result is that the research field has not yet achieved a mature state, and a variety of challenges still need to be overcome. The scaling to large size col-lections of documents, the use of more sophisticated natural language processing techniques for generating abstracts, the availability of annotated summarization corpora for training and testing purposes, are some of these challenges.

This is also the case for the domain of medical documents. The study of existing summarization techniques in other domains, the examination of different types of medical documents and the study of techniques reported so far in the literature for medical summarization lead to certain interesting remarks concerning the promising paths for future research. These remarks are presented below in terms of the *summarization factors*.

In terms of the *input medium*, almost all methods concern summarization from text, although the specific domain can provide a lot of useful input in other media as well (e.g. speech, images, videos). Summarizing information from different media (e.g. spoken transcriptions and textual reports related to the specific echo-videos) is an important issue for practical applications, representing a promising path for future research and development.

Concerning the *number of the input documents*, both categories of techniques (single and multi-document) have been examined. As it is the case in other domains, apart from medicine, single-document summarization methods are mainly using extractive techniques, whereas almost all of the multi-document summarizers are based on abstractive techniques. However, the selection between the simpler extractive techniques and the more complex abstractive ones should not only be based on the number of input documents, but also on the available resources, tools and the *summary purpose and output factors*.

Concerning the language *of the input document(s)*, most of the existing systems are monolingual (English in almost all the cases). There are two cases (MiTAP, MUSI), where the multilingual aspect was taken into account. In the MiTAP case, this was due to the domain (monitoring disease outbreaks) where the information sources are in various languages. On the other hand, MUSI summarizes the articles of a bilingual Journal. In the medical domain there is an enormous amount of documents in various categories (e.g. patient records) in other languages apart from English. There are resources and tools in several other languages that can be exploited in building summarizers for handling more than one language using either shallow or deeper approaches for language processing.

In relation to *purpose factors*, the existing methods mainly concern indicative summarization. The purpose of such summaries is to navigate the reader to the required information, which seems to be sufficient for most practical applications in medicine as long as no better solution is available. The production of indicative summaries seems to "indicate" that the shallow summarizing strategies used so far are not enough for producing informative or even critical summaries. Deeper language processing techniques [65] and their combination with shallow processing ones seems to be a promising path for future research in NLP in general and more specifically in summarization.

There is a trend towards *user-oriented summaries*, which is reasonable since summarization systems in the medical domain aim to cover the information needs of different user types (clinicians, researchers, patients) and specific users. User involvement does not concern only the submission of a query to the system, but also the summary customization and presentation according to the user's model. The PERSIVAL system [57] maintains information about the users' preferences taking into account their expertise in the domain, as well as the users' access tasks. The summary presentation can also be affected by the user's model (e.g. production of a summary in the form of hypertext, combination of text and images or video, etc.). Personalized access to medical information is a crucial issue and needs to be further investigated. There is a lot of expertise from the application of user modeling techniques in other domains which can also be exploited in the medical domain (see [66, 67]).

*Domain customization* is another significant issue. Most of the existing medical summarization systems are able to process documents belonging in specific sub-domains of medicine. Emphasis must be given to the development of technology that can be easily ported to new sub-domains. The development of open architecture systems with reusable and trainable components and resources is imperative in summarization technology. This is directly related to the ability of exploiting pre-existing medical knowledge resources. There are currently various knowledge repositories such as the Unified Medical Language System (UMLS)[19], and MeSH, which can be exploited in several ways by summarization engines. For instance, they can be used to locate interesting document(s), and interesting sentences inside those documents. They can even be used to create conceptual representations of the selected sentences in order to produce abstractive summaries in the same or in a different language. Such approaches are presented in the literature and can be further investigated. The development of customizable summarization technologies requires also in-depth study of the medical document types and medical sublanguage. A general-purpose system must be able to exploit the various characteristics of the medical documents. For instance, the sectioning of scientific articles, the specialized language used in emailed reports or in patient records are important features that can significantly affect the performance of the involved language processing tools. In general, the research community must cooperate towards the development of portable summarization technology and the medical domain can provide the necessary application areas.

---

[19] http://www.nlm.nih.gov/research/umls/

Concerning the *output factors*, the quality of the summarization output is strongly related to the summarization task. Therefore, qualitative and quantitative criteria need to be established following a study of the domain and the users' interests. In terms of the decision between extractive and abstractive techniques, as noted above, this has to take into account several factors related to the input documents, the purpose of the summary, the qualitative criteria established as well as the available resources and tools.

## 6. Conclusions

This survey presented the potential of summarization technology in the medical domain, based on the examination of the state of the art, as well as of existing medical document types and summarization applications.

The challenges that the summarization research has to overcome need to be viewed under the prism of the requirements of the specific field. The scaling to large collections of documents in various languages and from different media, the generation of informative summaries using more sophisticated language and knowledge engineering techniques, the generation of personalized summaries, the portability to new sub-domains, the design of evaluation scenarios which model real-world situations, the integration of summarization technology in practical applications such as the clinical workflow, are among the issues that the summarization community needs to focus on.

## Acknowledgements

The authors would like to thank the anonymous reviewers, as well as Dr. Constantine D. Spyropoulos and Dr. George Paliouras, for their helpful and constructive comments. Many thanks also to Ms Eleni Kapelou and Ms Irene Doura for checking the use of English.